# Transfer Representation Learning with TSK Fuzzy System

Peng Xu, Zhaohong Deng, *Senior Member, IEEE*, Jun Wang, *Member, IEEE*,
Qun Zhang, Shitong Wang

*Abstract*—Transfer learning can address the learning tasks of unlabeled data in the target domain by leveraging plenty of labeled data from a different but related source domain. A core issue in transfer learning is to learn a shared feature space in where the distributions of the data from two domains are matched. This learning process can be named as transfer representation learning (TRL). The feature transformation methods are crucial to ensure the success of TRL. The most commonly used feature transformation method in TRL is kernel-based nonlinear mapping to the high-dimensional space followed by linear dimensionality reduction. But the kernel functions are lack of interpretability and are difficult to be selected. To this end, the TSK fuzzy system (TSK-FS) is combined with transfer learning and a more intuitive and interpretable modeling method, called transfer representation learning with TSK-FS (TRL-TSK-FS) is proposed in this paper. Specifically, TRL-TSK-FS realizes TRL from two aspects. On one hand, the data in the source and target domains are transformed into the fuzzy feature space in which the distribution distance of the data between two domains is minimized. On the other hand, discriminant information and geometric properties of the data are preserved by linear discriminant analysis and principal component analysis. In addition, another advantage arises with the proposed method, that is, the nonlinear transformation is realized by constructing fuzzy mapping with the antecedent part of the TSK-FS instead of kernel functions which are difficult to be selected. Extensive experiments are conducted on the text and image datasets. The results obviously show the superiority of the proposed method.

## I. INTRODUCTION

When traditional machine learning algorithms are applied to various applications, including natural language processing [1], computer vision [2] and health informatics [3], ther are often confronted with two problems. On one hand, the labeled data are not always fully provided. On the other hand, the training data and test data may be collected from different distributions in real world applications. As an effective technique to solve the two problems, transfer learning has been widely studied in recent years. It can leverage the training data from the source domain to help the learning tasks in the target domain. Domain adaptation is an important paradigm in transfer learning research, which assumes that the data from both the source domain and the target domain are from different distributions but have the same task [4]. According to the availability of the labeled data in the target domain, domain adaptation can be categorized as semi-supervised domain adaptation and unsupervised domain adaptation. The former has a small amount of labeled data in the target domain whereas the latter has no labeled data in the target domain [5]. This paper focuses on the unsupervised domain adaptation that has fewer restrictions on the data.

Current researches on transfer learning mainly focus on feature-based transfer and model-based transfer. There are also studies exploring sample-based transfer [6], or the integration of these paradigms [7, 8]. The two mainstream paradigms are introduced below and their drawbacks are summarized accordingly.

Model-based transfer learning refers to transfer using the parametric relationship between the source domain and target domain. The model-based transfer learning methods often directly yield the classifier for the target domain. The typical characteristics of these algorithms are the parametric correlation hypothesis [9-11] or the parametric sharing hypothesis [12, 13]. The core idea of these methods is to add the parametric relationship to the training process of the classifier for the target domain. [9-11] are based on parametric correlation hypothesis and an auxiliary classifier on the source domain is used to help the classifier training on the target domain. [12, 13] are based on parametric sharing hypothesis and the distribution distance regularization term is added to the objective function of the classifier.

Feature-based transfer learning algorithms assume that there exists a shared feature space between the source domain and the target domain. And the distributions of the data from two domains are consistent in the shared feature space. The core idea of these methods is to learn such a feature space for the data from two domains and then arbitrary classifiers can be applied to the data in this feature space. To achieve the nonlinear feature transformation when learning the shared feature space, kernel functions are the most commonly used technique to transform the raw data into the high-dimensional space. Then the linear dimensionality reduction is adopted to map the data in the high-dimensional space to the low-dimensional shared feature space [8, 14-16]. There are mainly two ways to

This work was supported in part by the National Key Research Pro-gram of China under Grant 2016YFB0800803, the NSFC under Grant 61772239, the Jiangsu Province Outstanding Youth Fund under Grant BK20140001, the National First-Class Discipline Program of Light Industry Technology and Engineering under Grant LITE2018-02, and Basic Research Program of Jiangnan University Key Project in Social Sciences JUSRP1810ZD. (Corresponding author: Zhaohong Deng.)

P. Xu, Z. Deng, J. Wang, and S. Wang are with the School of Digital Media, Jiangnan University and Jiangsu Key Laboratory of Digital Design and Software Technology, Wuxi, China (e-mail: 6171610015@stu.jiangnan.edu.cn; dengzhaohong@jiangnan.edu.cn; wangjun_sytu@hotmail.com; wxwangst@aliyun.com).

Q. Zhang is with the Library, Jiangnan University, Wuxi, China (e-mail: 961044284@qq.com).

construct the shared feature space and they are reviewed in Section II.

The model-based and feature-based transfer learning methods have several drawbacks. On one hand, the performance of most model-based transfer learning algorithms heavily depends on the form of the selected classifier. On the other hand, the discriminant information and geometric properties of the data in feature-based transfer learning can be impaired during the feature transformation. There have been some work alleviating this problem [2, 17]. Besides, the kernel functions are lack of interpretability and it is thus difficult to select an appropriate kernel function to achieve nonlinear transformation.

As a more intuitive and interpretable modeling method, TSK fuzzy system (TSK-FS) has been applied to various applications [18-20]. It is a kind of data driven system composed of several IF-THEN fuzzy rules with high interpretability. There have been some transfer fuzzy systems proposed to provide better interpretability and deal with the uncertainty for transfer learning [21]. Based on TSK-FS, a model-based transfer learning method and its enhanced version are proposed in [22] and [23] respectively. They assume that the consequent parameters of TSK-FS between the source and target domains are correlated. A model-based transfer learning method with parametric sharing hypothesis for consequent parameters of TSK-FS is developed in [24]. In [25], a set of fuzzy rules of TSK-FS is firstly constructed on the source domain, and then feature space of the target domain is modified by learning a mapping to match the existing rules. As the extended work of [25], Hua et al. [26] proposed label space adaptation by learning another mapping in addition to input space adaptation. To improve the performance of transfer learning based on fuzzy systems, semi-supervised learning and active learning are also introduced into fuzzy transfer learning in [27] and [28]. These methods have achieved promising performance, but they are all model-based transfer learning methods and their performance heavily depends on the learning ability of the specific fuzzy systems.

To overcome the aforementioned drawbacks of the existing transfer learning algorithms, a feature-based transfer learning method based on TSK-FS, i.e., transfer representation learning with TSK-FS (TRL-TSK-FS) is proposed in this paper. TRL-TSK-FS treats a multi-output TSK-FS as the feature mapping and constructs a shared feature space for the data from two domains. The proposed method realizes transfer representation learning from two aspects. On one hand, the distribution distance of the data from two domains is minimized in the fuzzy feature space. On the other hand, the discriminant information and geometric properties are preserved by the forms the linear discriminant analysis (LDA) and principal component analysis (PCA).

TRL-TSK-FS effectively overcomes the aforementioned drawbacks and has the following advantages. Firstly, unlike the existing model-based transfer fuzzy systems, TRL-TSK-FS is a feature-based method. It is therefore more flexible to select classifiers and its performance will not be affected by the specific classifier. Secondly, TRL-TSK-FS leverages the discriminant information and preserves the geometric properties of the data by the forms of LDA and PCA during the feature transformation. Thirdly and most importantly, TRL-TSK-FS considers a multi-output TSK-FS as a feature transformation method which realizes nonlinear transformations and linear dimensionality reduction simultaneously. Unlike the classical transfer representation learning methods which adopted kernel methods to realize nonlinear transformation [8, 14, 16, 17], the proposed method realizes nonlinear transformation by constructing the fuzzy mapping with the antecedent part of TSK-FS. The fuzzy mapping not only avoids the selection of kernel functions, but also preserves more information in the original data. The consequent part of TSK fuzzy system can be regarded as the linear dimensionality reduction which maps the data from high-dimensional space generated by the antecedents to the low-dimensional shared feature space.

The main contributions of the paper are summarized as follows:

1) The paper firstly introduces the TSK-FS into the feature-based transfer learning, and proposed a novel TRL-TSK-FS method to learn a shared feature space in which the distribution distance and the information loss of the data are minimized simultaneously.
2) A novel method for calculating the antecedent parameters of TSK-FS is proposed. It is based on a deterministic clustering algorithm Var-Part, and it avoids the initialization sensitivity problem in the traditional clustering based TSK-FS modeling method.
3) Extensive experiments are conducted on the image dataset Office-Caltech and text dataset NG20. The experimental results obviously show the effectiveness and superiority of the proposed method.

The remainder of the paper is organized as follows. Some classical transfer representation learning methods are reviewed first in Section II, followed by the fundamentals of the proposed method. The details of the proposed method are illustrated in Section III. In Section IV, the experimental results and the analysis are given. The conclusions are represented in Section V.

## II. RELATED WORK

The existing feature-based transfer learning algorithms are reviewed first in this section. And then the fundamentals of the proposed methods, including TKS fuzzy system and maximum mean discrepancy are described briefly.

### A. Transfer Representation Learning

The feature-based transfer learning is also known as transfer representation learning. There are mainly two ways to construct the shared subspace in current researches of transfer representation learning. One is hidden subspace constructing and the other is subspace alignment. Pan et al. [29] formulated minimizing distribution distance between two domains in the new shared subspace as the problem of solving kernel matrix. A nonlinear mapping is obtained implicitly by solving kernel matrix with semidefinite programing. Linear dimensional reduction was then conducted on the data mapped with kernel matrix and the hidden subspace was obtained. But the semidefinite programing has high computational complexity and is not suitable for large-scale data. As the improvement of [29], the transfer component analysis (TCA) proposed in [14] directly minimized the distribution distance of the data of two domains in the hidden subspace constructed by kernel function mapping and linear dimensional reduction. TCA formulated optimization problems as the generalized eigenvalue decomposition. As the extension of TCA, Long et al [16] proposed joint domain adaptation (JDA) which minimized the distribution distance of two domains not only in marginal distribution but also in conditional distribution. Integrating sample-based transfer and feature-based transfer, transfer joint matching (TJM) was proposed in [8].To preserve the geometric properties of the original data in the process of feature transformation, scatter component analysis (SCA) is proposed in [17]. There are two problems when constructing the hidden subspace in the above algorithms: 1) there may not be such a hidden subspace, and 2) it is difficult to select an appropriate kernel function to achieve nonlinear transformation. For the first problem, subspace alignment has been widely used in transfer representation learning to deal with it. The subspace alignment (SA) algorithm proposed in [30] first conducted dimensional reduction for the source domain and the target domain using principal component analysis and then the linear mapping from the source domain to the target domain is solved. [31] introduced kernel technique into the SA to realize the nonlinear transformation. For the second problem, the multiple kernel learning method was introduced into transfer learning to alleviate it in [32]. In this study, the proposed method TRL-TSK-FS will explicitly constructs a nonlinear transformation through fuzzy mapping to solve this problem.

*B. TSK fuzzy system*

TSK fuzzy system (TSK FS) [33] is an intelligent model based on fuzzy logic [34]. It can learn the model parameters by data-driven way like other machine learning methods [35]. TSK-FS has good interpretability and can be formulated as "IF-THEN" rules as follows:

$$\begin{aligned} &\text{IF: } x_1 \text{ is } A_1^k \wedge x_2 \text{ is } A_2^k \wedge \cdots \wedge x_d \text{ is } A_d^k, \\ &\text{THEN: } f^k(\boldsymbol{x}) = p_0^k + p_1^k x_1 + \cdots + p_d^k x_d, \\ &k = 1, 2, \cdots, K. \end{aligned} \quad (1)$$

where $k = 1, 2, \cdots, K$, the TSK-FS has $K$ rules. $\mathbf{x} \in R^{d \times 1}$, $d$ is the dimension of samples. $f^k(\mathbf{x})$ represents the output of the $k$th rule of TSK-FS. $A_i^k$ represents a fuzzy set. Unlike the crisp set where the membership can only be 0 or 1, the membership in fuzzy set can be any values between 0 and 1. The membership can be calculated using membership function. So the definition of membership functions is the core issue in TSK-FS.

According to different application scenarios, different forms of membership functions can be defined. In the absence of domain knowledge, a commonly used fuzzy membership function is the Gaussian function as in (2) [36]. Each fuzzy set has a corresponding membership function and the parameters $c_i^k, \delta_i^k$ in the membership function are called antecedent parameters. Fuzzy c-means is often used to calculate antecedent parameters [24].

$$\mu_{A_i^k}(x_i) = \exp\left(\frac{-(x_i - c_i^k)^2}{2\delta_i^k}\right) \quad (2)$$

With known antecedent parameters, the membership value of each feature to the corresponding fuzzy set $A_i^k$ can be calculated by (2). If the multiplication is used as conjunction operator, the fire level of the $k$th rule of each sample can be calculated as (3) and its normalized form is as that in (4). The output of the TSK-FS is the weighted average of $f^k(\mathbf{x})$ as that in (5).

$$\mu^k(\mathbf{x}) = \prod_{i=1}^{d} \mu_{A_i^k}(x_i) \quad (3)$$

$$\tilde{\mu}^k(\mathbf{x}) = \mu^k(\mathbf{x}) \bigg/ \sum_{k'=1}^{K} \mu^{k'}(\mathbf{x}) \quad (4)$$

$$f(\mathbf{x}) = \sum_{k=1}^{K} \tilde{\mu}^k(\mathbf{x}) f^k(x) \quad (5)$$

Once the antecedent parameters are obtained, the output of TSK-FS in (5) can be re-expressed in (6) [37].

$$y = f(\mathbf{x}) = \mathbf{p}_g^T \mathbf{x}_g \quad (6)$$

where:

$$\mathbf{x}_e = [1, \mathbf{x}^T]^T \in R^{(d+1) \times 1} \quad (7a)$$

$$\tilde{\mathbf{x}}^k = \tilde{\mu}^k(\mathbf{x}) \mathbf{x}_e \in R^{(d+1) \times 1} \quad (7b)$$

$$\mathbf{x}_g = [(\tilde{\mathbf{x}}^1)^T, (\tilde{\mathbf{x}}^2)^T, \cdots, (\tilde{\mathbf{x}}^K)^T]^T \in R^{K(d+1) \times 1} \quad (7c)$$

$$\mathbf{p}^k = [p_0^k, p_1^k, \cdots, p_d^k]^T \in R^{(d+1) \times 1} \quad (7d)$$

$$\mathbf{p}_g = [(\mathbf{p}^1)^T, (\mathbf{p}^2)^T, \cdots, (\mathbf{p}^K)^T]^T \in R^{K(d+1) \times 1} \quad (7e)$$

where $\mathbf{x}_g$ represents the feature vector through fuzzy mapping of the antecedent part of TSK-FS and $\mathbf{p}_g$ represents the consequent parameters of the TSK-FS. In literatures, (6) is viewed as a linear model and $\mathbf{p}_g$ can be solved by least square method with the known $\mathbf{x}_g$.

*C. Maximum Mean Discrepancy*

Transfer representation learning assumes that the distribution distance between two domains is minimized in the new feature space. Maximum mean discrepancy (MMD) is a commonly used distribution distance measure in transfer learning [14, 38]. MMD is a two-sample statistical test method. According to the distribution distance of the observed data $\mathbf{X} = \{\mathbf{x}_i\}_{i=1}^{m}$ and $\mathbf{Y} = \{\mathbf{y}_j\}_{j=1}^{n}$ from distribution $p$ and $q$ respectively, MMD can reject or accept the null hypothesis $p = q$. In practice, the empirical estimation of MMD can be formulated as follows in (8) [39].

$$\text{MMD}^2(\mathbf{X}, \mathbf{Y}) = \left\| \frac{1}{m} \sum_{i=1}^{m} \phi(\mathbf{x}_i) - \frac{1}{n} \sum_{j=1}^{n} \phi(\mathbf{y}_j) \right\|_{\mathcal{H}}^2 \quad (8)$$

where $\phi$ is the feature mapping and $\|\cdot\|_{\mathcal{H}}$ represents the Reproducing Kernel Hilbert Space. Minimizing (8) means that minimizing the distribution distance between the source domain and the target domain under the mapping $\phi$. The goal of the transfer learning is to solve $\phi$.

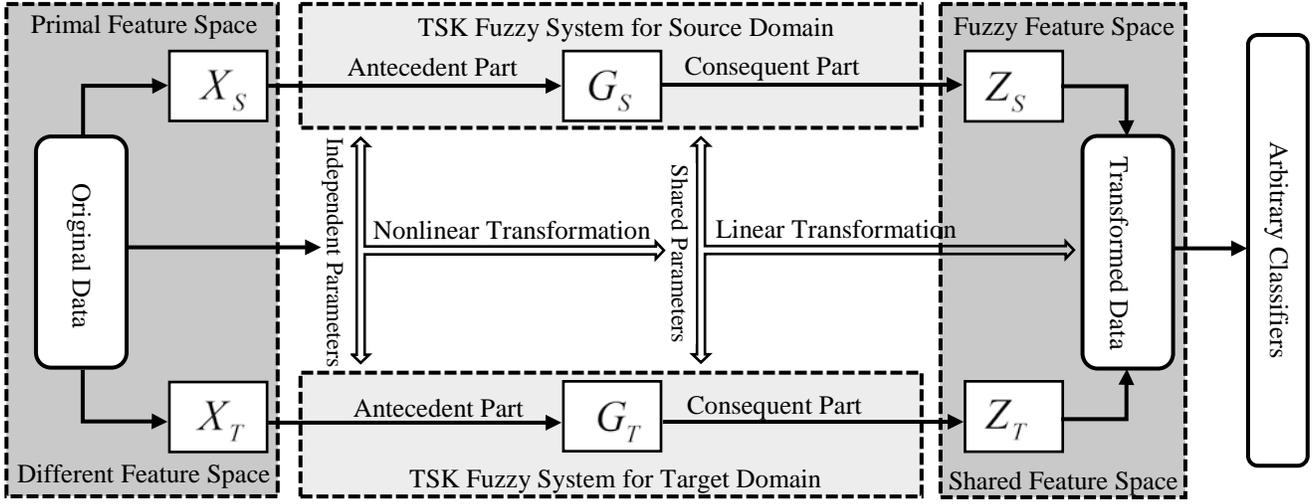

Fig. 1 The flowchart of the proposed method. The figure can be analyzed through three parts: the gray, light graph and white parts. The gray part represents the procedure of transformation of the feature space. The light gray part represents two TSK fuzzy systems. And the white part represents the data flow in the whole procedure. Through the transformation, the generated data can be fed into any classifiers.

In unsupervised domain adaptation, given the data and labels $\mathbf{X}_S = \{\mathbf{x}_{s_i}\}_{i=1}^{n_s}$, $\mathbf{Y}_S = \{y_{s_i}\}_{i=1}^{n_s}$ in the source domain and data with pseudo labels $\mathbf{X}_T = \{x_{t_j}\}_{j=1}^{n_t}$, $\hat{\mathbf{Y}}_T = \{\hat{y}_{t_j}\}_{j=1}^{n_t}$ in the target domain. Let $P_S(\mathbf{X}_S)$ and $P_T(\mathbf{X}_T)$ represent the marginal distributions of two domains, $Q_S(\mathbf{Y}_S | \mathbf{X}_S)$ and $Q_T(\mathbf{Y}_T | \mathbf{X}_T)$ represent the conditional distributions of two domains. To simultaneously adapt the marginal distributions and conditional distributions of two domains, the MMD can be defined as (9a) and (9b) respectively [16].

$$\mathrm{MMD}_P^2(\mathbf{X}_S, \mathbf{X}_T) = \left\| \frac{1}{n_s} \sum_{i=1}^{n_s} \phi(\mathbf{x}_{s_i}) - \frac{1}{n_t} \sum_{j=1}^{n_t} \phi(\mathbf{x}_{t_j}) \right\|_{\mathcal{H}}^2 \quad (9a)$$

$$\mathrm{MMD}_Q^2(\mathbf{X}_S, \mathbf{X}_T) = \sum_{c=1}^{C} \left\| \frac{1}{n_s^{(c)}} \sum_{y_{s_i}=c} \phi(\mathbf{x}_{s_i}) - \frac{1}{n_t^{(c)}} \sum_{\hat{y}_{t_j}=c} \phi(\mathbf{x}_{t_j}) \right\|_{\mathcal{H}}^2 \quad (9b)$$

where $C$ is the number the classes. $n_s^{(c)}$ represents the number of examples belonging to the $c$th class in the source domain and $n_t^{(c)}$ represents the number of examples belonging to the $c$th class in the target domain. The initial pseudo labels for the target domain are obtained by the classifier trained on the source domain and the accuracy is low. Iterative strategy is adopted to improve the accuracy of pseudo labels, where each time the joint distributions are matched and then the classifier is trained again to label the data in the target domain.

## III. TRANSFER REPRESENTATION LEARNING WITH TSK-FS

### A. Overview of RTL-TSK-FS

The shared feature space construction is fundamental in transfer representation learning. During the procedure of feature transformation, the constraints for transfer learning are added into the objective function. There are two crucial constraints for transfer representation learning. One is distribution matching where the distribution distance of the two domains is minimized. The other is preserving discriminant information and geometric properties. Based on above discussion, the problem can be expressed as (10).

$$\min_{\phi} Distance(\mathbf{X}_S, \mathbf{X}_T | \phi) + Info\_loss(\mathbf{X}_S, \mathbf{X}_T | \phi) \quad (10)$$

The first term represents distribution matching under the mapping $\phi$. The purpose of the second term is to minimize the information loss which contains discriminant information and geometric properties. $\phi$ can be learning by minimizing (10). In the existing methods, the most commonly used form of $\phi$ is kernel mapping followed by linear dimensionality reduction. The form of $\phi$ in RTL-TSK-FS is the TSK-FS.

The rationales of the proposed RTL-TSK-FS are described through five parts as follows. The construction method of the shared feature space through TKS-FS is described in detail in Section III-B. The two constraints for transfer learning are represented in Section III-C and Section III-D respectively. The optimization of the overall objective function and analysis of computational complexity are given in Section III-E and Section-F.

### B. Shared Feature Space Construction

It is fundamental to construct the shared feature space in transfer representation learning. In general, there are two steps to construct the feature space. The first step is nonlinear transformation and the second step is linear dimensionality reduction.

The proposed method realizes non-linear transformation by the antecedent part of the multi-output TSK-FS and linear dimensionality reduction by the consequent part of the multi-output TSK-FS. The flow chart of the transformation procedure is illustrated in Fig.1.

#### 1) Fuzzy feature space based on TSK-FS

Take a multi-output TKS-FS as a feature transformation $\phi$, for an example $\mathbf{x}_{s_i}$ in the source domain or an example $\mathbf{x}_{t_i}$ in the target domain, the transformed data $\mathbf{g}_{s_i}$ and $\mathbf{g}_{t_i}$ through the antecedent part based on (7c) can be represented as (11a) and 11(b).

$$\mathbf{g}_{s_i} = [(\tilde{\mathbf{x}}_{s_i}^1)^T, (\tilde{\mathbf{x}}_{s_i}^2)^T, \cdots, (\tilde{\mathbf{x}}_{s_i}^K)^T]^T \in R^{K(d+1) \times 1} \quad (11a)$$

$$\mathbf{g}_{t_i} = [(\tilde{\mathbf{x}}_{t_i}^1)^T, (\tilde{\mathbf{x}}_{t_i}^2)^T, \cdots, (\tilde{\mathbf{x}}_{t_i}^K)^T]^T \in R^{K(d+1) \times 1} \quad (11b)$$

$$\mathbf{G}_S = [\mathbf{g}_{s_1}, \mathbf{g}_{s_2}, \cdots, \mathbf{g}_{s_{n_s}}] \in R^{K(d+1) \times n_s} \quad (11c)$$

$$\mathbf{G}_T = [\mathbf{g}_{t_1}, \mathbf{g}_{t_2}, \cdots, \mathbf{g}_{t_{n_t}}] \in R^{K(d+1) \times n_t} \quad (11d)$$

$$\mathbf{P} = [\mathbf{p}_g^1, \mathbf{p}_g^2, \cdots, \mathbf{p}_g^m] \in R^{K(d+1) \times m} \quad (11e)$$

where $\mathbf{G}_S$ in (11c) is the concatenated data of $\mathbf{g}_{s_i}$ for all examples in the source domain and $\mathbf{G}_T$ in (11d) can be obtained in a similar way. The main distinction between the multi-output TSK-FS and the single-output TSK-FS as in (6) is that there are multiple-group consequent parameters in the multi-output TSK-FS. $\mathbf{P}$ in (11e) is the consequent parameter for the $m$-dimensional output TSK-FS. Then for an example $\mathbf{x}_{s_i}$ or $\mathbf{x}_{t_i}$ in the source domain or the target domain, the transformed data can be represented as (12a) and (12b).

$$\phi(\mathbf{x}_{s_i}) = \mathbf{P}^T \mathbf{g}_{s_i} \quad (12a)$$

$$\phi(\mathbf{x}_{t_i}) = \mathbf{P}^T \mathbf{g}_{t_i} \quad (12b)$$

By the transformation of the $m$-dimensional output TSK-FS, all of the data of the source domain and the target domain in the fuzzy feature space can be represented as the matrix forms in (13a) and (13b) respectively.

$$\mathbf{Z}_S = \mathbf{P}^T \mathbf{G}_S \quad (13a)$$

$$\mathbf{Z}_S = \mathbf{P}^T \mathbf{G}_S \quad (13b)$$

When the multi-output TSK-FS is regarded as the feature mapping function, two TSK fuzzy systems are applied on the data of the source domain and the target domain respectively as the illustration in Fig.1. In the paper, we assume that the consequent parameters of two TSK-FSs are shared, while the antecedent parameters are different. Take the antecedent parameters of TSK-FS for the source domain as $\mathbf{C}_S \in R^{K \times d}$ and $\mathbf{D}_S \in R^{K \times d}$, and that for the target domain as $\mathbf{C}_T \in R^{K \times d}$ and $\mathbf{D}_T \in R^{K \times d}$. The $\mathbf{C}_S$ and $\mathbf{C}_T$ represent the center of the membership function in (2). The $\mathbf{D}_S$ and $\mathbf{D}_T$ represent the kernel width of the membership function in (2). The consequent parameters are solved in Section III-E. And the antecedent parameters for both domains are calculated in advance with unsupervised clustering as following.

*2) Calculation of antecedent parameters*

Fuzzy c-means (FCM) is a common method to obtain the antecedent parameters of TSK-FS, but its stability is poor due to the random initialization of FCM. For the algorithms which need parameter optimization, the FCM based TSK-FS is sensitive to the parameters and the practicability is reduced. In the proposed method TRL-TSK-FS, a deterministic clustering algorithm Var-Part [40] is adopted to obtain the antecedent parameters. First, Var-Part algorithm is used to cluster data from the source domain and the target domain respectively and two center matrices $\mathbf{C}_S$ and $\mathbf{C}_T$ can be obtained. $K$ is the number of clusters and is also the number of rules of TSK-FS.

To cluster the data using Var-Part algorithm, for the selected cluster $C_j$, it is to firstly compute the variance in each dimension and find the dimension with the largest variance, such as $d_p$. Then, let $x_{ip}$ denote the value of example $\mathbf{x}_i$ in feature $d_p$, $\mu_{jp}$ denote the mean of $C_j$ in feature $d_p$, and divide $C_j$ into two sub-clusters $C_{j1}$ and $C_{j2}$, according to the following rule: If $x_{ip}$ is less than or equal to $\mu_{jp}$, assign $\mathbf{x}_i$ to $C_{j1}$; otherwise, assign $\mathbf{x}_i$ to $C_{j2}$. The above process completes one partition, that is, one cluster becomes two clusters. Then choose the next cluster to partition by selecting the cluster with the largest within-cluster sum-squared-error and repeat the above process until $K$ clusters are produced.

Once $\mathbf{C}_S$ and $\mathbf{C}_T$ are known, the kernel width matrices $\mathbf{D}_S$ and $\mathbf{D}_T$ are calculated as (14) in the similar way with FCM in [37]. And then the kernel width for each dimension is scaled to the range [1, 10], which is a reasonable range based on extensive experiments.

$$(\mathbf{D}_S)_p^k = \sum_{i=1}^{n_s}(x_{s_{ip}} - (\mathbf{C}_S)_p^k)^2 \quad (14a)$$

$$(\mathbf{D}_T)_p^k = \sum_{i=1}^{n_t}(x_{t_{ip}} - (\mathbf{C}_T)_p^k)^2 \quad (14b)$$

where $k = 1, 2, \cdots, K$ and $K$ is the number of fuzzy rules, $d$ is the dimension of the examples with $p = 1, 2, \cdots, d$. (14a) and (14b) are used for calculating the kernel widths in the source domain and the target domain, respectively, where $x_{s_{ip}}$ and $x_{t_{ip}}$ represent the value of examples in the source domain and the target domain in feature $d_p$.

*C. Distribution Matching*

One of two crucial aspects for transfer representation learning is distribution matching. The most commonly used method MMD is adopted in the paper to match the distributions for the source and target domains in the fuzzy feature space. Based on (9a), (12a) and (12b), the empirical MMD of the marginal distributions between two domains in the fuzzy feature space can be expressed as (15a).

$$\mathrm{MMD}_P^2(\mathbf{Z}_S, \mathbf{Z}_T) = \left\| \frac{1}{n_s} \sum_{i=1}^{n_s} \mathbf{P}^T \mathbf{g}_{s_i} - \frac{1}{n_t} \sum_{j=1}^{n_t} \mathbf{P}^T \mathbf{g}_{t_j} \right\|_{\mathcal{H}}^2 \quad (15a)$$
$$= \mathrm{Tr}(\mathbf{P}^T \mathbf{G}_X \mathbf{M} \mathbf{G}_X^T \mathbf{P})$$

where $\mathbf{G}_X = [\mathbf{G}_S, \mathbf{G}_T] \in R^{K(d+1) \times (n_s + n_t)}$, $\mathbf{M} \in R^{(n_s + n_t) \times (n_s + n_t)}$. The specific form of $\mathbf{M}$ is as follows.

$$\mathbf{M}_{ij} = \begin{cases} \dfrac{1}{n_s^2}, & i, j \leq n_s \\ \dfrac{1}{n_t^2}, & i, j \leq n_s \\ -\dfrac{1}{n_s n_t}, & \text{otherwise} \end{cases} \quad (15b)$$

Minimizing the distribution distance in the marginal distribution does not guarantee that the distribution distance in the conditional distributions is also minimized. To match the joint distributions between two domains, conditional distribution matching should also be considered and the empirical MMD can be expressed as (16a) based on (9b), (12a) and (12b).

$$\text{MMD}_Q^2(\mathbf{Z}_S, \mathbf{Z}_T) = \sum_{c=1}^{C} \left\| \frac{1}{n_s^{(c)}} \sum_{y_{s_i}=c} \mathbf{P}^\text{T} \mathbf{g}_{s_i} - \frac{1}{n_t^{(c)}} \sum_{\hat{y}_{t_i}=c} \mathbf{P}^\text{T} \mathbf{g}_{t_i} \right\|_{\mathcal{H}}^2 \quad (16a)$$
$$= \text{Tr}(\mathbf{P}^\text{T} \mathbf{G}_X \sum_{c=1}^{C} \mathbf{M}_c \mathbf{G}_X^\text{T} \mathbf{P})$$

where $\mathbf{M}_c \in R^{(n_s+n_t)\times(n_s+n_t)}$, $c = 1,2,\cdots,C$ and $C$ is the number of classes. The specific form of $\mathbf{M}_c$ is as follows. The pseudo labels $\hat{\mathbf{Y}}_T = \{\hat{y}_{t_j}\}_{j=1}^{n_t}$ are predicted by any classifier. The 1NN classifier is adopted in the paper.

$$(\mathbf{M}_c)_{ij} = \begin{cases} \dfrac{1}{n_s^{(c)} n_s^{(c)}}, & i,j \le n_s \text{ and } y_{s_i}, y_{s_j} = c \\ \dfrac{1}{n_t^{(c)} n_t^{(c)}}, & i,j > n_s \text{ and } \hat{y}_{t_{(i-n_s)}}, \hat{y}_{t_{(j-n_s)}} = c \\ -\dfrac{1}{n_s^{(c)} n_t^{(c)}}, & \begin{cases} i \le n_s, j > n_s \text{ and } \hat{y}_{s_i}, \hat{y}_{t_{(j-n_s)}} = c \\ i > n_s, j \le n_s \text{ and } \hat{y}_{t_{(i-n_s)}}, \hat{y}_{s_j} = c \end{cases} \\ 0, & \text{otherwise} \end{cases} \quad (16b)$$

(15a) and (16a) correspond to the first term of (10). Although (10) minimizes the objective function with respect to feature mapping $\phi$, $\phi$ can be decomposed into non-linear transformation and linear transformation. The non-linear transformation can be achieved by the antecedent part of TSK-FS with unsupervised clustering. So the optimizable parameters are only the linear mapping $\mathbf{P}$, that is, the consequent parameters of TSK-FS. Minimizing the distribution difference in the fuzzy feature space can be re-expressed as (17).

$$\min_{\mathbf{P}} \text{Tr}(\mathbf{P}^\text{T} \mathbf{G}_X \mathbf{M} \mathbf{G}_X^\text{T} \mathbf{P}) + \text{Tr}(\mathbf{P}^\text{T} \mathbf{G}_X \sum_{c=1}^{C} \mathbf{M}_c \mathbf{G}_X^\text{T} \mathbf{P}) \quad (17)$$

*D. Discriminant Information and Geometric Properties Preserving*

In addition to the distribution matching, it is also crucial to preserve the discriminant information and geometric properties which can be impaired during the procedure of distribution matching. If the data transformed by the antecedent part of the TSK-FS is viewed as the intermediate representations in the high-dimensional space, the consequent part of the TSK-FS can be viewed as linear dimensionality reduction on the intermediate representations. Then the discriminant information and geometric properties can be preserved during the optimization process of the consequent parameters of the TSK-FS.

Since the data in the target domain have no labels, the geometric properties can be preserved by maximizing the variance for the data in the fuzzy feature space like PCA. The optimization objective is formulated as follows.

$$\max_{\mathbf{P}} \text{Tr}(\mathbf{P}^\text{T} \mathbf{G}_T \mathbf{H}_T \mathbf{G}_T^\text{T} \mathbf{P}) \quad (18)$$

where $\mathbf{H}_T$ is the centering matrix which can centralize the examples so that the co-variance matrix can be directly computed as $\mathbf{G}_T \mathbf{H}_T \mathbf{G}_T^\text{T}$. $\mathbf{H}_T = \mathbf{I}_{n_t} - (1/n_t)\mathbf{1}_{n_t}\mathbf{1}_{n_t}^\text{T}$, where $\mathbf{I}_{n_t}$ is the identity matrix and $\mathbf{1}_{n_t}$ is the column vector with all ones. Here $n_t$ is the number of examples in the target domain and represents the dimension of $\mathbf{H}_T$.

For the data in the source domain, there are labels for the examples. Discriminant information should be preserved and the optimization objective can be formalized like LDA, that is, maximizing the between-class scatter and minimizing the within-class scatter in the fuzzy feature space. The optimization objective is as follows.

$$\max_{\mathbf{P}} \frac{\text{Tr}(\mathbf{P}^\text{T} \mathbf{S}_b \mathbf{P})}{\text{Tr}(\mathbf{P}^\text{T} \mathbf{S}_w \mathbf{P})} \quad (19)$$

where $\mathbf{S}_b$ is between-class scatter matrix and $\mathbf{S}_w$ is within-class scatter matrix, the specific forms of which are as follow.

$$\mathbf{S}_b = \sum_{c=1}^{C} n_s^{(c)} (\mathbf{m}_s^{(c)} - \bar{\mathbf{m}}_s)(\mathbf{m}_s^{(c)} - \bar{\mathbf{m}}_s)^\text{T} \quad (20)$$

$$\mathbf{S}_w = \sum_{c=1}^{C} \mathbf{G}_S^{(c)} \mathbf{H}_S^{(c)} (\mathbf{G}_S^{(c)})^\text{T} \quad (21)$$

where $\mathbf{m}_s^{(c)}$ is the mean of data $\mathbf{G}_S$ belonging to the class $c$ in the source domain. $\bar{\mathbf{m}}_s$ is the mean of data $\mathbf{G}_S$ belonging to all classes in the source domain. The definition of $\mathbf{H}_S^{(c)}$ is similar to $\mathbf{H}_T$ in (18) with replacement of $n_t$ using $n_s^{(c)}$, and the $n_s^{(c)}$ is the number of examples belonging to the class $c$ in the source domain.

Considering the discriminant information and geometric properties simultaneously, minimizing the information loss in the fuzzy feature space can be re-expressed as (22) which is corresponding to the second term in (10).

$$\min_{\mathbf{P}} \frac{\text{Tr}(\mathbf{P}^\text{T} \mathbf{S}_w \mathbf{P})}{\text{Tr}(\mathbf{P}^\text{T} \mathbf{G}_T \mathbf{H}_T \mathbf{G}_T^\text{T} \mathbf{P}) + \text{Tr}(\mathbf{P}^\text{T} \mathbf{S}_b \mathbf{P})} \quad (22)$$

*E. Overall Objective Function and Optimization*

The problem defined in transfer representation learning in (10) can be tackled with integrating (18) and (22). By introducing $\text{Tr}(\mathbf{P}^\text{T}\mathbf{P})$ minimization to avoid overfitting and regularization parameters for each term, the overall objective function can be formulated as follows.

$$\min_{\phi} Distance(\mathbf{X}_S, \mathbf{X}_T | \phi) + Info\_loss(\mathbf{X}_S, \mathbf{X}_T | \phi)$$
$$= \min_{\mathbf{P}} \frac{\text{Tr}(\mathbf{P}^\text{T} \mathbf{G}_X (\mathbf{M} + \sum_{c=1}^{C} \mathbf{M}_c) \mathbf{G}_X^\text{T} \mathbf{P}) + \alpha \text{Tr}(\mathbf{P}^\text{T}\mathbf{P}) + \beta \text{Tr}(\mathbf{P}^\text{T} \mathbf{S}_w \mathbf{P})}{\lambda \text{Tr}(\mathbf{P}^\text{T} \mathbf{G}_T \mathbf{H}_T \mathbf{G}_T^\text{T} \mathbf{P}) + \beta \text{Tr}(\mathbf{P}^\text{T} \mathbf{S}_b \mathbf{P})}$$
$$= \min_{\mathbf{P}} \frac{\text{Tr}(\mathbf{P}^\text{T} (\mathbf{G}_X (\mathbf{M} + \sum_{c=1}^{C} \mathbf{M}_c) \mathbf{G}_X^\text{T} + \alpha \mathbf{I} + \beta \mathbf{S}_w) \mathbf{P})}{\text{Tr}(\mathbf{P}^\text{T} (\lambda \mathbf{G}_T \mathbf{H}_T \mathbf{G}_T^\text{T} + \beta \mathbf{S}_b) \mathbf{P})}$$

(23)

where $\alpha$, $\beta$ and $\lambda$ are the regularization parameters for the 2-norm regularization term, discriminant information preserving term and geometric properties preserving term, respectively. The consequent parameters in $\mathbf{P}$ can be obtained by solv-ing (23). Observing that scaling **P** does not influence the results of (23), the objective function can thus be re-formulated as (24).

$$\min_P \text{Tr}(\mathbf{P}^T(\mathbf{G}_X(\mathbf{M}+\sum_{c=1}^C \mathbf{M}_c)\mathbf{G}_X^T + \alpha \mathbf{I} + \beta \mathbf{S}_w)\mathbf{P}) \quad (24)$$
$$\text{s.t. } \text{Tr}(\mathbf{P}^T(\lambda \mathbf{G}_T \mathbf{H}_T \mathbf{G}_T^T + \beta \mathbf{S}_b)\mathbf{P}) = 1$$

where $I \in R^{K(d+1) \times K(d+1)}$ is the identity matrix. (24) can be optimized with the Lagrange function as (25).

$$L = \text{Tr}(\mathbf{P}^T(\mathbf{G}_X(\mathbf{M}+\sum_{c=1}^C \mathbf{M}_c)\mathbf{G}_X^T + \alpha \mathbf{I} + \beta \mathbf{S}_w)\mathbf{P})$$
$$+ \text{Tr}((\mathbf{P}^T(\lambda \mathbf{G}_T \mathbf{H}_T \mathbf{G}_T^T + \beta \mathbf{S}_b)\mathbf{P} - \mathbf{I})\Phi) \quad (25)$$

where $\Phi = \text{diag}(\varphi_1, \varphi_2, \cdots, \varphi_m) \in R^{m \times m}$ is Lagrangian multiplier. By Setting $\frac{\partial L}{\partial \mathbf{P}} = 0$, the following equation is obtained.

$$(\mathbf{P}^T(\mathbf{G}_X(\mathbf{M}+\sum_{c=1}^C \mathbf{M}_c)\mathbf{G}_X^T + \alpha \mathbf{I} + \beta \mathbf{S}_w)\mathbf{P})$$
$$= (\lambda \mathbf{G}_T \mathbf{H}_T \mathbf{G}_T^T + \beta \mathbf{S}_b)\mathbf{P}\Phi \quad (26)$$

Based on (26), the optimization problem in (23) is transformed into the problem of generalized eigenvalue decomposition. Finding the optimal **P** is then reduced to solve (26) for the $m$ smallest eigenvectors. $\varphi_1, \varphi_2, \cdots, \varphi_m$ are the $m$ smallest eigenvalues and $\mathbf{P} = [\mathbf{p}_g^1, \mathbf{p}_g^2, \cdots, \mathbf{p}_g^m]$ are the corresponding eigenvectors. Once the consequent parameters **P** are obtained, the new generated data $\mathbf{Z}_S$ and $\mathbf{Z}_T$ in the fuzzy feature space can be obtained easily.

After getting the new representations through transfer learning, arbitrary classifiers can be applied to classify the data in the target domain. Due to the existence of joint distribution adaptation, the process of optimization is iterative. Each time the new representations are obtained, the classifier is used to update the labels of the data in the target domain, and then the above process is repeated. The description of the algorithm TRL-TSK-FS is summarized in Table 1.

TABLE I
DESCRIPTION OF TRL-TSK-FS ALGORITHM

**Algorithm**: RTL-TSK-FS
**Input**: Data in the source domain $\mathbf{X}_S = \{\mathbf{x}_{s_i}\}_{i=1}^{n_s}$ and the corresponding labels $\mathbf{Y}_S = \{y_{s_i}\}_{i=1}^{n_s}$; the data in the target domain $\mathbf{X}_T = \{x_{t_i}\}_{i=1}^{n_t}$; trade-off parameters $\alpha$, $\beta$ and $\lambda$; the number of fuzzy rules $K$; the dimension of fuzzy feature space $m$; the number of iterations for joint distribution adaptation $T$; the classifier used to train the new data and label the data in target domain.
**Output**: New data in the fuzzy feature space $\mathbf{Z}_S$ and $\mathbf{Z}_T$.
**Procedure MV-ITCC**:
1: Calculate the antecedent parameters $\mathbf{C}_S$, $\mathbf{C}_T$, $\mathbf{D}_S$ and $\mathbf{D}_T$ for the data $\mathbf{X}_S$ and $\mathbf{X}_T$ based on algorithm Var-Part and (12).
2: Calculate $\mathbf{G}_S$ and $\mathbf{G}_T$ using $\mathbf{C}_S$, $\mathbf{C}_T$, $\mathbf{D}_S$ and $\mathbf{D}_T$ based on 7(a) – 7(c).
3: **For** $t \leftarrow 1, 2, \ldots, T$ **do**
4: Update $\mathbf{M}$ and $\mathbf{M}_c$ based on (14) and (16).
5: Update $\mathbf{S}_b$ and $\mathbf{S}_w$ based on (20) and (21)
6: Update $\mathbf{P}$ based on (25) using generalized eigenvalue decomposition.
7: Update $\mathbf{Z}_S$ and $\mathbf{Z}_T$ based on (11a) and (11b).
8: Train the selected classifier using $\mathbf{Z}_S$ and $\mathbf{Z}_T$, and update the pseudo labels for the data in the target domain $\hat{\mathbf{Y}}_T = \{\hat{y}_{t_i}\}_{i=1}^{n_t}$.
9: **end for**

### F. Computational Complexity

Based on the above discussion, we will analyze the efficiency of the proposed method. The computational complexity of Algorithm RTL-TSK-FS in Table I is analyzed using the big $O$ notation. Denote the number of examples as $N$ and $N = n_s + n_t$, the dimension of the original data as $d$ and the number of fuzzy rules as $K$. For the step 1, the computational complexity for calculating the $\mathbf{C}_S$, $\mathbf{C}_T$, $\mathbf{D}_S$ and $\mathbf{D}_T$ is $O(2dNK)$. For the step 2, the computational complexity for calculating the $\mathbf{G}_S$ and $\mathbf{G}_T$ the computational complexity is $O(2dNK+2NK)$ based on (2), (3), (4) and (7b). Denote the iterations for joint distribution adaptation as $T$. The computational complexity of step 4 for constructing MMD matrices is $O(TCN^2)$ with the number of classes as $C$. In step 5, the computational complexity for $\mathbf{S}_b$ and $\mathbf{S}_w$ is $O(TCN + TNK^2d^2 + TN^2Kd)$. Denote the dimension of fuzzy feature space as $m$, the computational complexity of generalized eigenvalue decomposition for step 6 is $O(Tmk^2d^2)$. The computational complexity of constructing the data $\mathbf{Z}_S$ and $\mathbf{Z}_T$ for step 7 is $O(TNmd)$. Based on the above analysis, the major computation cost lies in the MMD matrices construction in step 4, matrix multiplication in step 5 and generalized eigenvalue decomposition in step 6. The computational cost can increase exponentially with the increase of size the examples. With the increase of number of fuzzy rules, the computational cost also increases exponentially and can be intensive, especially faced with high-dimensional data. Set $a = \max(T, C, m)$ and $b = \max(N, Kd)$, $a \ll b$ in general. Hence, the maximum overall computational cost is $O((4a^2 + 2ab)b^2)$.

## IV. EXPERIMENTS

### A. Datasets

The experiments are conducted to evaluate the effectiveness of the proposed algorithm on the commonly used image and text datasets for transfer learning. They are the image transfer dataset Office-Caltech and text transfer dataset NG20.

The Office-Caltech dataset composes of Office dataset and Caltech-256 dataset. Office [41, 42] is a benchmark dataset widely used for transfer learning, which contains 31 different categories. The images of Office dataset come from three domains which are AMAZON (images downloaded from online website, that is, www.amazon.com), Webcam (low resolution images by a simple Web camera) and DSLR (high resolution images by a digital SLR camera). Caltech-256 [43] is a famous object recognition dataset which contains 256 different categories. In the experiments, the same settings are used as

[16]. A (AMAZON), W (Webcam), D (DSLR) and C (Caltech-256) are treated as four domains and ten categories included in all four domains are used for the experiments. The SURF [44] feature of the selected images is extracted and the codebook is constructed by K-means to represent the images as the form of 800-bins with bag of visual work model. Finally, all data are standardized. Arbitrary two domains are selected from four domains as the source domain and the target domain and 12 distinct transfer tasks can be constructed from these four domains, e.g., C→A, C→W, …, D→W.

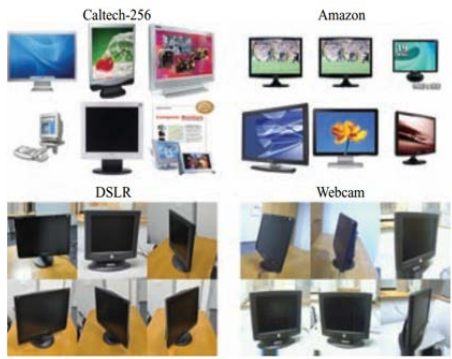

Fig.1. Example images of Office-Caltech dataset

TABLE II
THE CONSTRUCTION OF NG20

|  | Domain-A | Domain-B |
|---|---|---|
| COMP | graphics<br>os.ms-windows.misc | sys.ibm.pc.hardware<br>sys.mac.hardware |
| REC | autos<br>motorcycles | sport.baseball<br>sport.hockey |
| SCI | crypt<br>electronics | med<br>space |
| TALK | politics.guns<br>politics.mideast | politics.misc<br>religion.misc |

NG20 is a text transfer dataset constructed from 20-Newsgroup which contains more than 20000 examples belonging to six major categories and 20 sub-categories [45, 46]. In the experiments, four major categories are selected, which are denoted as COMP, REC, SCI and TAKL. Table II shows the details of the selection of sub-categories. Arbitrary two major categories are selected to a transfer task and 12 transfer tasks are constructed on the selected documents. For example, when COMP and REC are selected, the Domain-A of them are combined as the data in the source domain and the Domain-B of them are combined as the data in the target domain. Then this transfer task is represented as COMP→REC. In the same way, if the Domain-A is treated as the data in the target domain and Domain-B is treated as the data in the source domain, the transfer task is represented as REC→COMP. Then 12 transfer tasks can be constructed from four major categories. In order to improve the efficiency of the algorithm, deep learning method Doc2vec [47] was used to extract the features for all selected documents. This method can be used to represent every document with different length as the vectors with fixed length in the form of unsupervised learning. Before using the Doc2vec model to train document vectors, word segmentation, stop words removing, stemming and lemmatization operations are adopted to preprocess the documents [48]. The documents vectors with 200 dimensions are extracted finally.

*B. Experimental Settings*

In the experiments, two benchmark algorithms and five transfer representation learning methods are used as the comparing algorithms. Two benchmark algorithms are 1 nearest neighbor (1NN) and traditional TSK-FS. 1NN is applied on the original data and data after dimensionality reduction with principal component analysis. For TSK-FS, the dimension of new representations is high with the increasing of number of rules, then ill-posed problem may arise when the least square method is used to solve the consequent parameters of TSK-FS. Ridge regression technique is adopted to alleviate the above problem for TSK-FS [49]. Five transfer representation learning algorithms are respectively transfer component analysis (TCA), geodesic flow kernel (GFK) [42], joint distribution adaptation (JDA), transfer joint matching (TJM) and scatter component analysis (SCA). 1NN is used to classify the new representations obtained by these five algorithms.

Because there are no labels for the data in the target domain, cross-validation cannot be used to find the optimal parameters for all comparing algorithms. Thus all methods are evaluated by empirically searching the parameter space for the optimal parameter settings and only the best results are reported for each method. For the number of rules TSK-FS, the parameter is optimally set by searching grid $k=\{3,5,10\}$ For five transfer representation learning methods, the dimension of subspace is optimally set by searching grid $m=\{10,20,\cdots,100\}$. Except for special declaration, all trade-off parameters and regularization parameters existed in all algorithms are optimally set by searching grid $\lambda=\{0.01,0.1,1,10,100\}$, For all algorithms involving joint distribution adaptation, the number of iterations is set as $T=5$ because pseudo labels need several iterations to improve the model accuracy. For the trade-off parameter of SCA which controls the balance of the total scatter and the between-class scatter, it is optimally set by $\beta=\{0.1,0.3,0.5,0.8,1\}$

For the proposed method TRL-TSK-FS, the trade-off parameters $\alpha$, $\beta$, $\lambda$ and the number of iterations for joint distribution adaptation are optimally set by the same settings like the other comparing algorithms. Because the deterministic clustering algorithm is adopted to calculate the antecedent part of the TSK-FS, there is no parameter to be optimized unlike the commonly used method fuzzy c-means. An important parameter in TRL-TSK-FS is the number of rules which can influence the dimension of new representations in the fuzzy feature space and the effect of non-linear transformation for domain adaptation. It is optimally set by searching grid $K=\{3,4,5,\cdots,10\}$.

*C. Results Analysis*

*1) On image dataset*

Table III illustrates the accuracy of all algorithms on image transfer dataset Office-Caltech. Office-Caltech is a widely used benchmark dataset in transfer representation learning. In Table III, except for TSK and TRL-TSK-FS methods, the accuracy of 1NN (raw), 1NN (PCA), TCA and GFK are reported as results in [8], and the other algorithms are reported as the results in the corresponding original papers. For the algorithms involving kernel functions, the reported results have considered linear kernel function and radial basis kernel function. The method TSK (raw) is following the training strategy of traditional machine learning methods which treat the data in the source domain as the training data and the data in the target domain as the test data. TRL-TSK ($K$=3) denotes the accuracy of the proposed method TRL-TSK-FS when the number of rule is 3. The optimal results for each transfer task in Table III are highlighted in bold. It can be observed that the accuracy of TRL-TSK-FS is superior to all the other algorithms on Office-Caltech dataset. The accuracy of TJM is the highest in all comparing algorithms considering the average accuracy of 12 transfer tasks. The average accuracy of the proposed method is nearly higher 8% than that of TJM under of rule number of 3. The average accuracy of the proposed method is higher about 23% than that of 1NN (raw).

*2) On text dataset with linear kernel function*

Table IV illustrates the accuracy of all algorithms on text transfer dataset NG20. Since the NG20 dataset is constructed in the paper, the results of all algorithms are obtained according to the experimental settings in Section IV-B. Linear kernel function is selected for the algorithms involving kernel functions. The optimal results for each transfer task in Table IV are highlighted in bold. The training strategy of methods 1NN (raw), 1NN (PCA) and TSK (raw) is the same with TSK (raw) on the image dataset. The method 1NN (raw) classifies the original data directly while the method 1NN (PCA) classifies the data after dimensionality reduction using PCA. It is obvious that the average accuracy of TRL-TSK-FS on 12 transfer tasks outperforms all the other algorithms on NG20 dataset. The average accuracy of the proposed method is nearly higher 2% than that of JDA which achieved best results in all comparing algorithms. The first two algorithms 1NN (raw) and 1NN (PCA) have the lowest average accuracy in all algorithms. However, the accuracy of 1NN classifier increased by about 7% after PCA dimensionality reduction. The average accuracy of TSK (raw) are higher than that of 1NN (raw). It is because that the learning ability of TSK-FS exceeds that of 1NN. It is unreasonable to make direct comparison between TSK-FS and the other algorithms. Even so, the performance of TRL-TSK-FS with 1NN classifier is still superior to TSK-FS classifier, which verifies the effectiveness of transfer representation learning.

TABLE III
ACCURACY ON IMAGE TRANSFER DATASET OFFICE-CALTECH (%)

| Transfer Tasks | 1NN (raw) | 1NN (PCA) | TSK (raw) | TCA | GFK | JDA | TJM | SCA | TRL-TSK (K=3) |
|---|---|---|---|---|---|---|---|---|---|
| C→A | 23.70 | 36.95 | 52.40 | 45.82 | 41.02 | 44.78 | 46.76 | 43.74 | **58.46** |
| C→W | 25.76 | 32.54 | 47.46 | 30.51 | 40.68 | 41.69 | 38.98 | 33.56 | **49.83** |
| C→D | 25.48 | 38.22 | 44.59 | 35.67 | 38.85 | 45.22 | 44.59 | 39.49 | **46.50** |
| A→C | 26.00 | 34.73 | 43.99 | 40.04 | 40.25 | 39.36 | 39.45 | 38.29 | **45.06** |
| A→W | 29.83 | 35.59 | 38.98 | 35.25 | 38.98 | 37.97 | 42.03 | 33.90 | **50.85** |
| A→D | 25.48 | 27.39 | 43.95 | 34.39 | 36.31 | 39.49 | 45.22 | 34.21 | **46.50** |
| W→C | 19.86 | 26.36 | 33.84 | 29.92 | 30.72 | 31.17 | 30.19 | 30.63 | **38.56** |
| W→A | 22.96 | 29.35 | 37.68 | 28.81 | 29.75 | 32.78 | 29.96 | 30.48 | **45.62** |
| W→D | 59.24 | 77.07 | 82.80 | 85.99 | 80.89 | 89.17 | 89.17 | 92.36 | **94.27** |
| D→C | 26.27 | 29.65 | 30.81 | 32.06 | 30.28 | 31.52 | 31.43 | 32.32 | **36.24** |
| D→A | 28.50 | 32.05 | 33.82 | 31.42 | 32.05 | 33.09 | 32.78 | 33.72 | **45.30** |
| D→W | 63.39 | 75.93 | 81.36 | 86.44 | 75.59 | 89.49 | 85.42 | 88.81 | **94.24** |
| Average | 31.37 | 39.65 | 47.64 | 43.03 | 42.95 | 46.31 | 46.33 | 44.29 | **54.29** |

*3) On text dataset with radial basis function*

Table V shows the results of the algorithms involving kernel functions and the kernel function is set as radial basis function. It is still an open issue to select appropriate kernel functions for these algorithms. Besides linear kernel function, the radial basis function (RBF) is the commonly used nonlinear kernel function for feature transformation. Similar with that in [8, 17, 39], the form of RBF is $\exp(-\|\mathbf{a}-\mathbf{b}\|_2^2/\sigma^2)$ where the kernel width $\sigma$ is set to the median distance between samples in the aggregate domain in the experiments.

$$\sigma=\text{median}(\|\mathbf{a}-\mathbf{b}\|_2^2), \forall \mathbf{a},\mathbf{b} \in \mathbf{X}_S \cup \mathbf{X}_T \qquad (27)$$

Similar with the kernel functions, the antecedent part in TRL-TSK-FS also plays the role of nonlinear transformation. There are no parameters needed to be optimized in the antecedent part of TRL-TSK-FS. Unlike the kernel functions whose selection is blind and difficult, the fuzzy mapping of the antecedent part is more intuitive and interpretable. Let the other experimental settings be the same with that in previous experiments, the results of TCA, JDA, TJM and SCA are shown in Table V. It can be seen that the proposed method shows obvious superiority. The performance of SCA is poor with the RBF kernel function on the text dataset, which may be due to the reason that the selected kernel width is not ap-propriate for the text data. The above discussion verifies the effectiveness of TSK-FS for transfer representation learning.

TABLE IV
ACCURACY ON TEXT TRANSFER DATASET NG20 (%)

| Transfer Tasks | 1NN (raw) | 1NN (PCA) | TSK (raw) | TCA | GFK | JDA | TJM | SCA | TRL-TSK (K=3) |
|---|---|---|---|---|---|---|---|---|---|
| COM→REC | 57.21 | 69.32 | 86.06 | 81.74 | 86.87 | 92.76 | 90.45 | 90.43 | **97.44** |
| REC→COM | 59.99 | 61.00 | 68.43 | 82.66 | 73.96 | 93.61 | 90.92 | 93.81 | **94.04** |
| COM→SCI | 56.73 | 62.21 | 66.37 | 59.99 | 73.33 | 81.44 | 78.66 | 76.03 | **81.92** |
| SCI→COM | 53.72 | 57.44 | 62.75 | 64.23 | 67.71 | 75.49 | 73.66 | 75.64 | **75.92** |
| COM→TALK | 81.81 | 92.29 | 95.61 | 92.95 | 95.55 | 96.48 | 95.43 | 95.67 | **96.80** |
| TALK→COM | 86.27 | 92.52 | 95.38 | 94.98 | 95.06 | **96.43** | 96.24 | 95.43 | 95.90 |
| REC→SCI | 51.51 | 58.41 | 65.62 | 65.14 | 72.80 | 87.46 | 81.96 | **89.17** | 88.36 |
| SCI→REC | 58.50 | 58.95 | 63.47 | 65.82 | 67.91 | 85.84 | 85.43 | 80.79 | **87.38** |
| REC→TALK | 55.45 | 68.88 | 80.89 | 85.51 | 88.31 | 90.31 | 85.54 | 82.95 | **91.55** |
| TALK→REC | 71.30 | 72.13 | 81.31 | 90.62 | 83.73 | 90.80 | 91.32 | 89.36 | **92.81** |
| SCI→TALK | 62.51 | 76.15 | 80.77 | 78.52 | 79.79 | 78.88 | 79.02 | 76.24 | **84.64** |
| TALK→SCI | 69.93 | 72.42 | 80.34 | 84.08 | 80.13 | 82.80 | 84.86 | 85.86 | **86.88** |
| Average | 63.74 | 70.14 | 77.25 | 78.85 | 80.43 | 87.69 | 86.12 | 85.95 | **89.47** |

TABLE V
ACCURACY ON TEXT TRANSFER DATASET NG20 (%)

| Transfer Tasks | TCA | JDA | TJM | SCA | TRL-TSK |
|---|---|---|---|---|---|
| COM→REC | 81.18 | 93.17 | 87.66 | 87.46 | **97.44** |
| REC→COM | 81.57 | 93.28 | 90.44 | 84.56 | **94.04** |
| COM→SCI | 59.15 | 79.70 | 76.16 | 62.52 | **81.92** |
| SCI→COM | 63.51 | 75.62 | 74.09 | 67.00 | **75.92** |
| COM→TALK | 92.35 | 96.21 | 95.22 | 94.21 | **96.80** |
| TALK→COM | 95.40 | **96.43** | 96.09 | 94.20 | 95.90 |
| REC→SCI | 64.86 | 87.15 | 79.19 | 73.35 | **88.36** |
| SCI→REC | 64.91 | 86.57 | 85.38 | 56.43 | **87.38** |
| REC→TALK | 84.57 | 89.84 | 83.19 | 78.21 | **91.55** |
| TALK→REC | 90.22 | 92.18 | 90.67 | 88.24 | **92.81** |
| SCI→TALK | 78.58 | 80.56 | 77.63 | 78.70 | **84.64** |
| TALK→SCI | 83.74 | 82.77 | 85.18 | 79.08 | **86.88** |
| Average | 78.34 | 87.79 | 85.08 | 78.66 | **89.47** |

### D. Parameters Analysis

*1) The number of fuzzy rules*

Fig.2 illustrates the average accuracy under different number of rules of TSK-FS on the image and text datasets. According to the analysis of computational complexity, with the increase of fuzzy rule numbers, the computational complexity can increase exponentially. The maximum number of rules is set as 10 in the experiments. The accuracy on the image dataset is shifted on the whole for the sake of more clear display, and it does not affect the analysis for the variation tendency. For the image dataset Office-Caltech, the promising accuracy can be obtained under the number of 5, 6 and 7. Meanwhile, the accuracy will decrease with the increasing number when the number exceeds 7. For the image dataset NG20, the accuracy keeps rising with the increase of rules. The dimension for the image dataset is 800, while the dimension for the text dataset is 200. With the increase of rules, the parameters and complexity of the model for the image dataset will increase rapidly. There may be overfitting for the image dataset when the number of rules exceeds 7. So it can be concluded from Fig.2 that a few rules are enough for the dataset with high dimension and increasing number of rules can result in steady improvement for the dataset with low dimension.

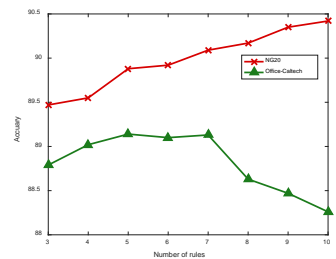

Fig.2. Accuracy with different number of rules

*2) The number of iterations for joint distribution adaptation*

Fig.3 (a) and Fig.3 (b) illustrate the accuracy under different number of iterations for each task on the image and text datasets, respectively. For the sake of clarity, the accuracy for each task is moved up or down on the whole, which does not affect the trend analysis. From Fig.3 (a) on the image dataset, it can be seen that the proposed method has good convergence on the 12 tasks of the image dataset. The accuracy is fairly stable when the number of iterations exceeds 5. From Fig.3 (b) on the text dataset, we can see that most tasks have good convergence after 5 iterations. Only the accuracy of task C→W is fluctuant under different iterations, while the amplitude of fluctuation decreases as the number of iterations increases. Therefore, it is reasonable to set the number of iterations to 5 in the experiments.

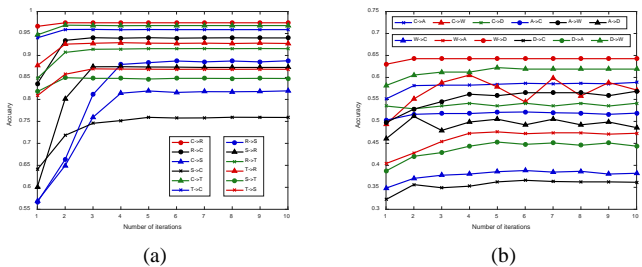

(a)     (b)

Fig.3. Convergence analysis on image and text dataset.

*3) The dimensionality of fuzzy feature space*

The dimensionality of fuzzy feature space is treated as *m* according to the *m*-dimensional output TSK-FS. Fig.4 (a) and Fig.4 (b) illustrate the accuracy under different dimensionality of fuzzy feature space for each task on the image and text dataset. Only four representative tasks are selected for both datasets to observe the variation tendency more clearly. From Fig.4 (a) on the image dataset, the dimensionality of the fuzzy feature space of the optimal accuracy varies under different transfer tasks. Although four tasks in Fig.4 (a) show different trends in classification accuracy as the dimensionality increases, the overall trend is relatively stable and the optimal accuracy can be obtained within 100 dimensions. From Fig.4 (b) on the text dataset, the classification accuracy of four tasks shows downward trends with the increase of dimensionality. It can be seen that on the text dataset, the low dimensional feature space can get better classification accuracy.

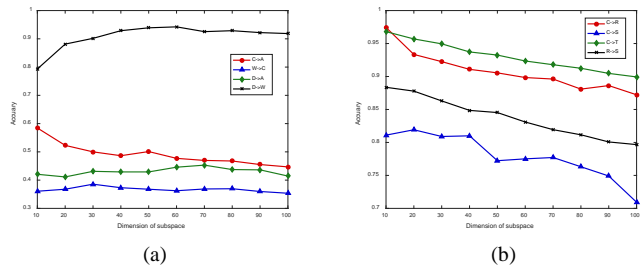

(a)     (b)

Fig.4 Dimension analysis.

*4) Trade-off parameters*

The trade-off parameters $\alpha$, $\beta$ and $\lambda$ are analyzed based on the experimental results. Take six transfer tasks on the image dataset as the standard, Fig.5 (a) – Fig.5 (f) show the effect of parameters on classification accuracy under different tasks. Fig.5 (a) and Fig.5 (b) show the effect of parameter α on the classification accuracy with the other parameters fixed. It can be seen that α = 0.1 can get promising results in most cases. Similarly, Fig.5 (c) and Fig.4 (d) indicate that $\beta$=0.01 performs better than the other values. It can be concluded from Fig.5 (e) and Fig.5 (f) that $\lambda$=0.01 or $\lambda$=0.1 can always achieve better results. In conclusion, 0.1 and 0.01 are the more reliable selections for these three parameters under all tasks.

## V. CONCLUSIONS

A novel transfer representation learning method RTL-TSK-FS based on TSK-FS is proposed in this study. From the aspect of TSK-FS, unlike the other fuzzy transfer learning methods where TSK-FS is treated as a classifier, TSK-FS is regarded as a feature learning method in the proposed method. From the aspect of feature learning, unlike the traditional methods using kernel functions, RTL-TSK-FS provides a new idea for non-linear transformation by the way of fuzzy mapping. There are various methods to construct the fuzzy mapping and a deterministic clustering method is adopted in the paper to avoid the problem of initialization sensitivity. The results of experiments on the image and text datasets show that the proposed method is superior to some state-of-art transfer representation learning methods.

In the future, we will make in-depth research in transfer representation learning based on TSK-FS from the following aspects. In the proposed method RTL-TSK-FS, the consequent parts of the TSK-FS of two domains are shared. Therefore, RTL-TSK-FS is only appropriate for the homogeneous transfer learning and cannot be applied to the scene of heterogeneous transfer learning [50]. Further researches for the heterogeneous transfer learning will be explored in the future to enhance the transfer learning ability of the TSK-FS. Furthermore, PCA and LDA are used to preserve the data geometric properties in the proposed method, which can only preserve the global structures of the data. Other techniques that can preserve the local structures of the data, such as locality preserving projections (LPP) [51] will be studied in our future researches. In addition, how to improve the scalability of the proposed method is also a significant work in our future study.

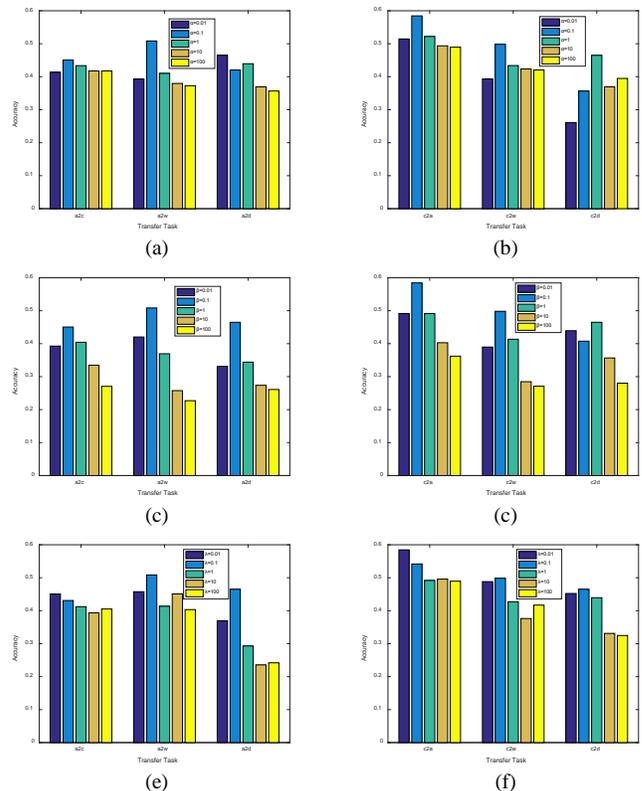

Fig.5 Trade-off parameters analysis.